\documentclass[10pt,twocolumn,letterpaper]{article}

\usepackage{cvpr}              
\usepackage{booktabs}
\usepackage{xcolor}
\usepackage{pifont}
\usepackage{makecell}
\usepackage{multirow}

\definecolor{cvprblue}{rgb}{0.21,0.49,0.74}
\usepackage[pagebackref,breaklinks,colorlinks,allcolors=cvprblue]{hyperref}
\newcommand{\cmark}{\textcolor{green}{\ding{51}}}
\newcommand{\xmark}{\textcolor{red}{\ding{55}}}

\def\paperID{*****} 
\def\confName{CVPR}  
\def\confYear{2026}

\title{FloodVLM-Pix: Language-Guided Pixel-Level Flood Understanding}

\author{
Chengyang Zhang\textsuperscript{1}, 
Zhihao Wang\textsuperscript{1}, 
Zhiyu Long\textsuperscript{1}, 
Xueqian Wang\textsuperscript{1}, 
Quanming Yao\textsuperscript{1}, 
Wei Wu\textsuperscript{2}, 
Gang Li\textsuperscript{1}\\
\textsuperscript{1}Tsinghua University, 
\textsuperscript{2}National Disaster Reduction Center}

\begin{document}
\maketitle
\begin{abstract}
As climate change intensifies, floods are becoming more frequent and destructive, making accurate flood understanding an urgent priority. However, existing vision-language models perform poorly on flood-related tasks, suffering from attentional bias and inaccurate descriptions of flood magnitude. To bridge this gap, we construct the first large-scale remote sensing flood multi-modal benchmark that supports language-level flood understanding and pixel-level flood segmentation. The benchmark contains 131,200 flood image patches and corresponding 443,100 textual descriptions for flood severity and evolution. Based on this benchmark, we propose FloodVLM-Pix, a language-guided pixel-level flood understanding model that includes a flood prior module and a multi-temporal flood pixel decoder to achieve accurate flood understanding. The flood prior module automatically distills sub-region flood spatial priors and inputs priors into FloodVLM-Pix, solving the attentional bias problem in flood language understanding. The flood pixel decoder introduces a novel general-specific feature fusion mechanism to address performance degradation caused by interference (e.g., clouds or mountain shadows) during flood segmentation. Experiments on FloodVL-Pix demonstrate that FloodVLM-Pix improves both flood description quality and pixel-level flood segmentation performance compared with state-of-the-art VLMs and flood segmentation models. Our code and data will be released at \href{https://github.com/Haomiaoyanmeng/FloodPix}{https://github.com/Haomiaoyanmeng/FloodPix}.
\end{abstract}

\section{Introduction}

\begin{figure}[!t]
    \centering
    \setlength{\abovecaptionskip}{-1pt}  
    \setlength{\belowcaptionskip}{-8pt}
    \includegraphics[width=0.95\columnwidth]{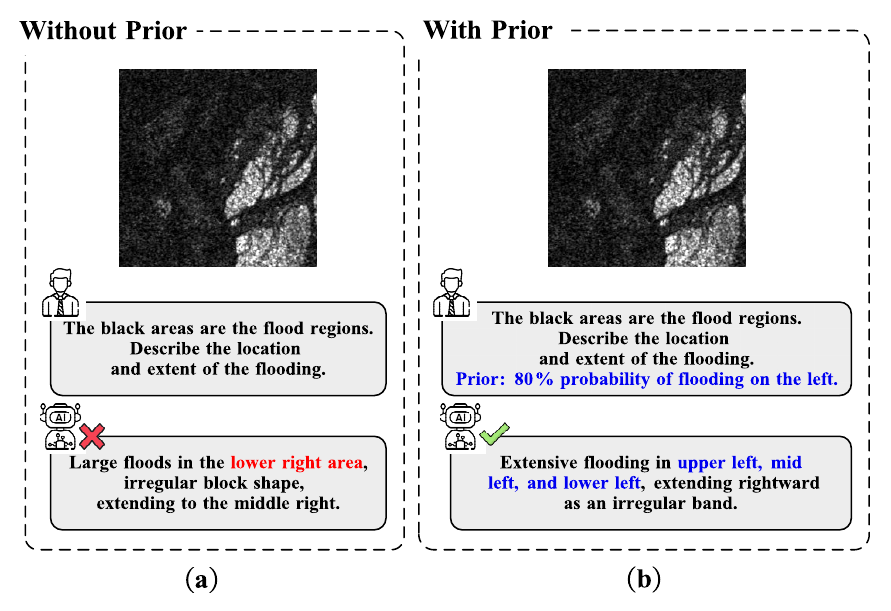}
    \caption{A SAR image patch with flood areas in the upper and left parts. Inference results of Qwen3.6-Plus \cite{qwen3}: (a) without prior, (b) with prior. In the absence of a prior (a), the Qwen3.6-Plus misinterprets large-scale flooding as background and thus ignores it. After incorporating the prior in (b), the Qwen3.6-Plus yields the correct result.}
    \label{fig:prior}
\end{figure}


Extreme flood events have become significantly frequent on a global scale \cite{flood-current, flood-global, flood-increasing, flood-mapping}, making flood understanding (including language-level and pixel-level analysis) from satellite remote sensing imagery increasingly important for disaster response and damage assessment \cite{dam}. Much of the existing remote sensing research on floods primarily focuses on pixel-level flood segmentation \cite{smagnet, flood-drought, flooddam2, floods1s2, list, dam, flood-continuous}. Flood segmentation localizes inundation but lacks the semantic and temporal abstraction required to characterize severity, progression, or impact of floods, limiting its utility for damage assessment and emergency response. Multi-temporal language-level flood analysis, in contrast, supplies this abstraction by translating visual evidence into descriptions that humans can rapidly interpret, enabling faster flood comprehension and more timely rescue action. Existing flood research remains flood-segmentation-centric and largely overlooks language-level flood understanding.

\begin{figure}[!t]
    \centering
    \setlength{\abovecaptionskip}{-1pt}  
    \setlength{\belowcaptionskip}{-7pt}
    \includegraphics[width=0.9\columnwidth]{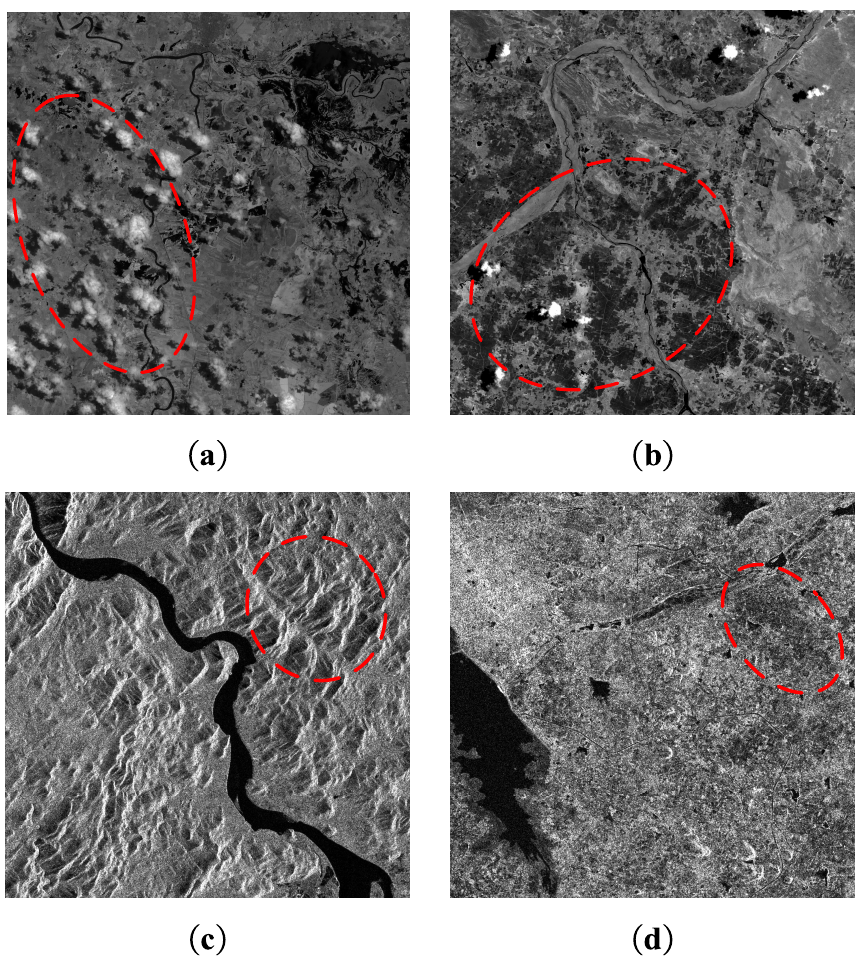}
    \caption{(a) Clouds and cloud shadows, (b) agriculture, (c) mountain shadows, and (d) wet surface areas in optical or SAR images, degrading the performance of flood segmentation. }
    \label{fig:flood_image_problem}
\end{figure}

Although existing vision-language models (VLM) can provide linguistic descriptions of images, they struggle with flood-related tasks, exhibiting inaccurate understanding of flood extent and severity. The limitations of VLMs arise from three factors. 1) The flood extent can be extremely large, sometimes covering over 80\% of a satellite image patch, inducing attentional bias that leads the model to misidentify the flood as background and describe non-flood areas instead, as shown in Fig. \ref{fig:prior}(a). 2) Existing VLMs are predominantly trained on massive optical datasets, resulting in the limited capability of these VLMs to interpret synthetic aperture radar (SAR) images \cite{earthgpt}. SAR imagery remains important for flood understanding due to its resistance to cloud cover and adverse weather conditions. 3) Flood images in each region are often taken at the same location but at different times and modalities. These images suffer from severe disturbances, including clouds, mountain shadows, agricultural areas, ambiguous wet regions, as shown in Fig. \ref{fig:flood_image_problem}.

To bridge the aforementioned gaps, we propose a benchmark, FloodVL-Pix, and a model, FloodVLM-Pix, for accurate flood understanding. Our contributions can be summarized as follows:
\begin{itemize}

    \item We construct FloodVL-Pix, the first flood-oriented remote sensing multi-modal multi-temporal benchmark that supports both language-level flood understanding and pixel-level flood segmentation. FloodVL-Pix comprises 106,500 SAR image patches and 24,700 optical image patches, among which 11,000 have flood coverage exceeding 70\%, providing abundant large-flood samples to mitigate attentional bias during optical and SAR flood interpretation. For researchers, FloodVL-Pix establishes a language-pixel-temporal paradigm for flood understanding, laying the foundation for next-generation disaster-oriented Earth observation VLMs that reason about disaster dynamics under real-world disturbances.

    \item We propose FloodVLM-Pix, a language-guided pixel-level flood understanding model, including \textbf{a flood prior module} and \textbf{a multi-temporal flood pixel decoder}. The prior module provides accurate flood priors to the VLM, addressing the attentional bias during flood image patch interpretation. The pixel decoder adaptively fuses general shared semantic features and specific per-image-patch features of multi-temporal flood image patches to solve the performance degradation caused by disturbances shown in Fig. \ref{fig:flood_image_problem}. For government and emergency agencies, FloodVLM-Pix delivers pixel-level inundation maps and language-level severity/evolution reports for flood assessment, evacuation, and relief logistics. For citizens in flood-prone regions, FloodVLM-Pix translates complex satellite imagery into accessible flood descriptions and flood severity information, bridging the last mile from raw observations to timely disaster response.

    
\end{itemize}

\section{Related Work}
\subsection{Flood Model/Benchmark}
As shown in Table~\ref{tab:benchmark_comparison}, existing remote sensing models and benchmarks designed for floods \cite{dam, sen12flood, flood-kuro, cauflood, floodnet} mostly focus on the flood segmentation task. This task is initially formulated for bi-temporal flood image inputs \cite{smagnet, flood-drought, flooddam2, floods1s2, list, dam, flood-sen1floods11}. For instance, \cite{smagnet} adaptively fuses SAR and multi-spectral images using a spatial mask and weight-shared decoder, achieving robust flood segmentation accuracy. Subsequently, some researchers begin to investigate flood segmentation of multi-temporal remote sensing flood images \cite{flood-kuro, flood-offshore, flood-continuous}. For example, \cite{flood-continuous} combines sparse Markov random field-based spatial clustering with temporal fusion of historical flood information to achieve robust continuous flood extent monitoring. However, none of these methods \cite{dam, flood-kuro, flood-offshore, flood-sen1floods11, smagnet, flood-drought, flooddam2, floods1s2, list, flood-continuous} considers language-level flood understanding, which is essential for conveying flood severity and evolution and enabling rapid disaster interpretation and response.

\begin{table*}[t]
\centering
\small
\renewcommand{\arraystretch}{1.0}
\setlength{\abovecaptionskip}{3pt}
\setlength{\belowcaptionskip}{-4pt}
\begin{tabular*}{\textwidth}{
  >{\centering\arraybackslash}m{3.3cm}  
  >{\centering\arraybackslash}m{3cm}  
  >{\centering\arraybackslash}m{2cm}  
  >{\centering\arraybackslash}m{2cm}  
  >{\centering\arraybackslash}m{2cm}  
  >{\centering\arraybackslash}m{2.2cm}  
}
\toprule

\noalign{\vspace{-5pt}}  
\multicolumn{2}{c}{\textbf{Models (Benchmarks)}} &
\makecell[c]{\textbf{Flood-Oriented}} &
\makecell[c]{\textbf{Language}} &
\makecell[c]{\textbf{Pixel Results}} &
\makecell[c]{\textbf{Multi-Temporal} \textbf{($\geq$ 3)}} \\ [-5pt]

\midrule

\multirow{5}{*}{\textbf{Flood Models}}
& S1GFloods\cite{dam} & \cmark & \xmark & \cmark & \xmark \\
& SEN12-FLOOD\cite{sen12flood} & \cmark & \xmark & \cmark & \cmark \\
& CAU-Flood\cite{cauflood} & \cmark & \xmark & \cmark & \xmark \\
& FloodNet\cite{floodnet} & \cmark & \xmark & \cmark & \xmark \\
& Kuro Siwo\cite{flood-kuro} & \cmark & \xmark & \cmark & \cmark \\

\midrule

\multirow{5}{*}{\makecell{\textbf{Vision-Language} \\ \textbf{Remote Sensing Models}}}
& RingMo-GPT\cite{ringmogpt} & \xmark & \cmark & \xmark & \xmark \\
& Earth-GPT\cite{earthgpt} & \xmark & \cmark & \xmark & \xmark \\
& GeoChat\cite{geochat} & \xmark & \cmark & \xmark & \xmark \\
& TeoChat\cite{teochat} & \xmark & \cmark & \xmark & \cmark \\
& UniRS\cite{unirs} & \xmark & \cmark & \xmark & \cmark \\

\midrule

\multirow{3}{*}{\makecell{\textbf{Vision-Language-Pixel} \\ \textbf{Remote Sensing Models}}}
& GeoPix\cite{geopix} & \xmark & \cmark & \cmark & \xmark \\
& Geopixel\cite{geopixel} & \xmark & \cmark & \cmark & \xmark \\
& SegEarth-R2\cite{segearth} & \xmark & \cmark & \cmark & \xmark \\

\midrule

\textbf{Proposed}
& FloodVLM-Pix & \cmark & \cmark & \cmark & \cmark \\

\bottomrule
\end{tabular*}

\caption{Comparison of different models/benchmarks in terms of flood-oriented design, language support, pixel-level labels, and multi-temporal analysis capability.}
\label{tab:benchmark_comparison}
\end{table*}

\subsection{Remote Sensing VLM/Benchmark}
Early works on large-scale remote sensing models primarily focus on constructing vision-only foundation models \cite{ringmo, anysat, omnisat, neural, mmearth, skysense}. These works typically collect ultra-large-scale unlabeled remote sensing images and perform self-supervised learning to train foundation models. Subsequently, attention shifts to VLMs for remote sensing images \cite{earthgpt, geochat, ringmogpt, skysensegpt, teochat}, which enables deep language-level understanding of remote sensing tasks. Table~\ref{tab:benchmark_comparison} shows the capability coverage of existing VLMs/benchmarks across flood-oriented design, language support, pixel-level outputs, and multi-temporal analysis. For example, TeoChat \cite{teochat} proposes a benchmark and a VLM for multi-temporal remote sensing data, enabling conversational reasoning over image sequences. To address the limitation that VLMs cannot output pixel-level results, \cite{geopixel, geopix} propose pixel-level benchmarks and VLMs that incorporate pixel decoders, maintaining language understanding while yielding high-quality pixel-level outputs. However, multi-temporal understanding remains unexplored in the framework of \cite{geopixel, geopix}, limiting their ability to analyze evolution in remote sensing images over time. Moreover, the above remote sensing VLMs/benchmarks are not specifically designed for floods, and they exhibit attentional bias when interpreting flood imagery, resulting in inaccurate language-level flood understanding.


\begin{figure}[!t]
    \centering
    \includegraphics[width=\columnwidth]{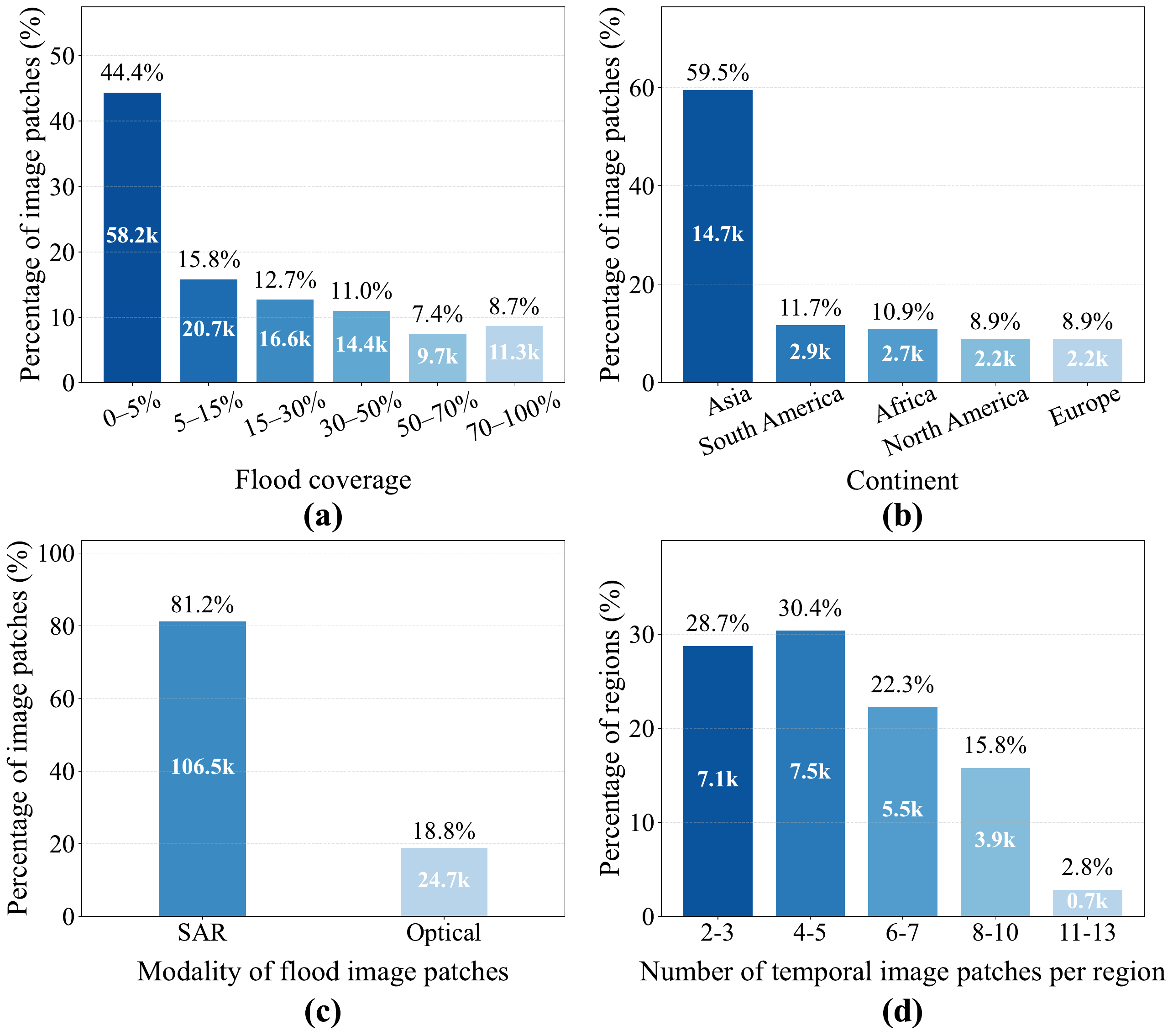}
    \setlength{\abovecaptionskip}{-6pt}
    \setlength{\belowcaptionskip}{-8pt}
    \caption{The statistics of the FloodVL-Pix dataset. (a) Flood coverage ratio distribution. (b) Continent distribution. (c) Modality distribution. (d) Temporal image patch number distribution per region. Numbers inside/above bars denote counts/percentage of flood image patches or regions. }  
    \label{fig:data_statistics}
\end{figure}

\section{Benchmark}

\noindent\textbf{Motivation.}
Optical remote sensing has been the mainstream data source for remote sensing VLMs, largely because large-scale image-text pretraining corpora are dominated by RGB optical imagery, whose visual cues are intuitive for cross-modal alignment. Floods are often accompanied by extreme precipitation and heavy cloud cover, making SAR images essential for reliable flood understanding due to the robustness of SAR imagery to adverse weather. However, existing VLMs struggle with SAR-based flood understanding due to mainly optical-image pretraining \cite{sarchat} and attentional bias toward overwhelmingly large flooded areas, as shown in Fig.~\ref{fig:prior}(a). To address these issues, we construct FloodVL-Pix, a dataset combining SAR and optical imagery to improve VLM performance on large-scale flood understanding in SAR and optical image patches.


\begin{figure*}[!h]
    \centering
    \includegraphics[width=\textwidth]{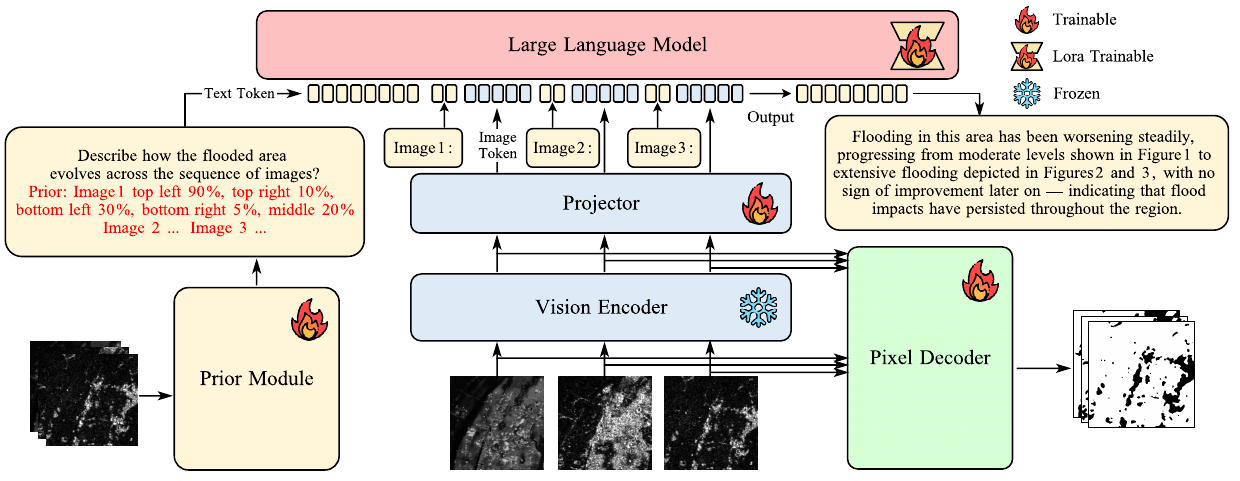}
    \setlength{\abovecaptionskip}{-3pt}  
    \setlength{\belowcaptionskip}{-7pt}
    \caption{Structure of our FloodVLM-Pix model. Our model consists of a vision encoder, a projector, an LLM, a prior module, and a pixel decoder. The vision encoder and projector convert image patches into tokens for LLM understanding. The prior module provides spatial priors for flood-related cues. The pixel decoder decodes image patch features to pixel-level flood segmentation results. The yellow and blue small rectangles below the LLM represent text tokens and visual tokens, respectively. ``Image 1/2/3'' denotes the order of input image patches.}  
    \label{fig:overview_archi}
\end{figure*}

\noindent\textbf{Dataset and statistics.}
FloodVL-Pix is built from 247 global flood scenes, each comprising 100 regions (finer scene), and each region contains $2\sim13$ multi-temporal, co-registered $256\times256$ flood image patches. We then annotate these patches with pixel-level flood segmentation labels and language descriptions. In total, FloodVL-Pix contains 131,200 image patches with pixel-level flood labels and corresponding 443,100 flood textual descriptions. Construction details are provided in the supplementary material.
We set the patch size to $256\times256$ for three main reasons. 1) $256\times256$ is a common patch size in flood segmentation benchmarks \cite{cauflood,dam,floodnet,worldfloods,ombria,unosat}, enabling fair comparison. 2) For transformer-based VLMs, full self-attention has both computational and memory complexities of $O(n^2)$ with respect to the token length $n$. Since $n \propto H \times W$ and $H=W$ in our work, the complexities become $O(H^4)$, so $256\times256$ patches substantially reduce computation and memory compared with large tiles such as $1024\times1024$. 3) Smaller patches yield finer-grained flood descriptions than large-tile interpretation, which benefits accurate flood response and rescue. 


Fig.~\ref{fig:data_statistics} summarizes the statistics. FloodVL-Pix contains 106,500 SAR and 24,700 optical image patches, of which 11,300 patches have $>70\%$ flood coverage, enabling VLM to effectively learn large-scale flood representations. Asia accounts for 59.5\% of the data, because the FloodVL-Pix includes data from flood-prone regions such as South/Southeast Asia and China. The $2\sim13$ temporal image patches per region support understanding of flood dynamics and evolution.


\section{Model}


\begin{figure*}[!t]
    \centering
    \includegraphics[width=\textwidth]{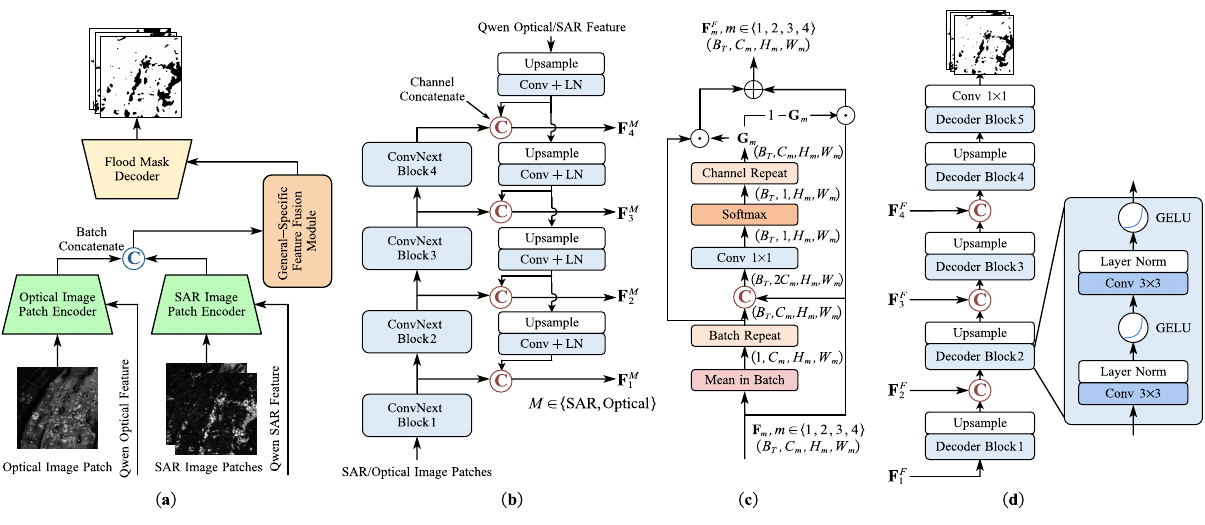}
    \setlength{\abovecaptionskip}{-7pt}  
    \setlength{\belowcaptionskip}{-7pt}
    \caption{Structures of our pixel decoder. (a) Overview of our pixel decoder. (b) Optical/SAR image patch encoder. (c) General-specific fusion module. (d) Flood mask decoder.} 
    \label{fig:pixel_deco}
\end{figure*}

\subsection{Overview of FloodVLM-Pix}
The overview architecture of FloodVLM-Pix is shown in Fig.~\ref{fig:overview_archi}. FloodVLM-Pix consists of 1) a prior module that supplies language-level flood priors to calibrate VLM attention towards floods, 2) a VLM backbone (vision encoder, projector, and LLM) that takes flood image patches and flood-oriented questions and priors as input and produces flood descriptions, and 3) a pixel decoder that fuses general and specific multi-temporal features for pixel-level flood segmentation. The prior module and VLM backbone are depicted in Section 4.2, and the pixel decoder is discussed in Section 4.3.


\subsection{Prior Module and VLM Backbone}


Existing VLMs exhibit attentional bias in remote sensing flood image patch [Fig.~\ref{fig:prior}(a)]. We argue this bias stems from a mismatch between the object-centric assumption of natural-image pretraining and the scene-centric nature of floods. Natural-image pretraining treats targets as compact, locally salient objects, while large homogeneous regions are absorbed into the background. Floods, however, are spatially extended states with weak boundaries and can cover over 80\% of an image patch. Consequently, attention is misallocated, treating flooded regions as background and underestimating flood extent. We therefore introduce sub-region-level flood extent priors, obtained by estimating sub-region flood ratios and verbalizing them as language-level evidence, to recalibrate attention toward flooded areas [Fig.~\ref{fig:prior}(b)] and improve flood severity and evolution-aware descriptions.

We first divide the flood image patch $\mathbf{I}^{H \times W \times 3}$ into five sub-regions: top-left ($0\sim H/2$, $0\sim W/2$), top-right ($0\sim H/2$, $W/2\sim W$), bottom-left ($H/2\sim H$, $0\sim W/2$), bottom-right ($H/2\sim H$, $W/2\sim W$), and center ($H/4\sim 3H/4$, $W/4\sim 3W/4$). $H$ and $W$ are the height and width of the image patch, and both are 256 in our work. These sub-regions $\mathbf{x} \in {\mathbb{R}^{B \times H/2 \times W/2 \times3}}$ are then fed into the prior module, where $B$ is the batch dimension. The prior module output $\hat{\mathbf{y}} \in {\mathbb{R}^{B \times 2}}$ representing the proportions of the sub-regions for flood occurrence and non-occurrence, respectively. For instance, an output value of [0.8, 0.2] for a sub-region indicates that the module estimates 80\% of the region as flooded and 20\% as non-flooded. The proportion label $\mathbf{y} \in {\mathbb{R}^{B \times 2}}$ is obtained by computing the flood/non‑flood area ratios in each sub-region from the flood segmentation labels. The prior module is trained using the mean squared error loss function i.e., 
\vspace{-0.0em}
\begin{equation}
\mathcal{L} = \frac{1}{B} \left\| \hat{\mathbf{y}} - \mathbf{y} \right\|_F^2
\end{equation}
\vspace{-0.0em}
to bring the predictions closer to the labels.

Once the prior module is trained, we freeze it to provide stable flood priors for the LLM, preventing subsequent fine-tuning from corrupting the learned flood-prior representations or causing catastrophic forgetting. We then convert the output proportions of the prior module for each sub-region into language-level flood priors. For example, a top-left sub-region with an output of 0.9 for flood occurrence is described as ``top left 90\%''. The priors contain accurate flood information for each sub-region in each temporal flood image patch, guiding the VLM to focus on flooded areas and mitigate attentional bias toward non-flooded regions, thereby improving VLM understanding of flood severity and evolution. 

Then, these flood priors, combined with flood-related questions as textual inputs, are used to fine-tune the LLM and projector to effectively learn flood representations from remote sensing flood image patches. Specifically, we initialize the vision encoder, projector, and LLM from Qwen3-VL-2B/4B \cite{qwen3}. To stabilize fine-tuning and reduce computational cost, we freeze the vision encoder and fine-tune the LLM using low-rank adaptation (LoRA) \cite{lora}. Meanwhile, we fine-tune all parameters of the projector to align flood representations across image and language modalities.



\subsection{Pixel Decoder}

The architecture of the pixel decoder is shown in Fig.~\ref{fig:pixel_deco}. Fig.~\ref{fig:pixel_deco}(a) shows the overview of the pixel decoder. To mitigate cross-modal interference during feature extraction, we employ separate optical and SAR encoders to independently extract modality-specific features from the optical and SAR flood image patches and the outputs from the Qwen3 vision encoder. Then the features from the optical and SAR encoders are concatenated in the batch dimension and input to the general-specific feature fusion module for deep-level multi-temporal feature fusion. Finally, the flood mask decoder decodes the fused features to the flood segmentation results. 

The image patch encoder is shown in Fig.~\ref{fig:pixel_deco}(b). We adopt ConvNeXt \cite{convnet} as the feature extractor for flood image patches due to its hierarchical design and large-kernel depthwise convolutions, which facilitate multi-scale spatial feature learning and fine-grained texture capture. The feature size from the Qwen3 vision encoder is $16 \times 16$. We gradually upsample and convolve the outputs of the Qwen3 vision encoder and concatenate them with the outputs of ConvNeXt at different scales to obtain hierarchical image patch features $\mathbf{F}^M_m$, where $M \in \{\mathrm{SAR}, \mathrm{optical}\}$ denotes the modality of the output features and $m \in \{1, 2, 3, 4\}$ denotes different scales. We then concatenate the optical and SAR features $\mathbf{F}^M_m$ ($M \in \{\mathrm{SAR}, \mathrm{optical}\}$) in a region (i.e., the features of all temporal image patches in a region) along the batch dimension to yield the concatenated features $\mathbf{F}_m \in \mathbb{R}^{B_T \times C_m \times H_m \times W_m}$, where $B_T$ equals the number of temporal image patches of a flood region, and $C_m$, $H_m$, and $W_m$ denote the number of channels, height, and width of the $\mathbf{F}_m$, respectively. The $\mathbf{F}_m$ contains the features from all modalities and temporal image patches in a region, enabling the general-specific module to sufficiently learn regional features.

The general-specific feature fusion module is the core of our pixel decoder, as shown in Fig.~\ref{fig:pixel_deco}(c). For image fusion in multi-temporal, multi-modal flood segmentation, the core challenge is that disturbances in flood images (i.e., clouds, shadows, wet soils, and irrigation) vary across modalities and times. These structured disturbances can mimic or obscure flood extent, requiring the fused representation to aggregate cross-modal and cross-temporal flood context under such disturbances while preserving image-specific information for pixel-level segmentation results. To address this challenge, we introduce general--specific feature fusion with two complementary feature spaces, where the general feature space captures region-level, cross-modal, cross-temporal flood features as robust contextual information, whereas the specific feature space preserves image-specific features for fine-grained flood texture delineation. Specifically, at each scale $m$, we treat the feature of the $i$-th temporal image patch, $\mathbf{F}_m^{(i)} \in \mathbb{R}^{1 \times C_m \times H_m \times W_m}, i\in\{1,2,...,B_T\}$, as a specific feature. These $B_T$ specific features are averaged over the batch dimension to uniformly incorporate features from all modalities and temporal image patches, which makes the averaged feature less affected by disturbances in any specific feature. The averaged feature is then repeated across the batch to obtain the general feature $\mathbf{F}_m^G$, where the batch repeat aligns the size of $\mathbf{F}_m^G$ with that of $\mathbf{F}_m$ for subsequent concatenation and fusion. We then concatenate $\mathbf{F}_m^G$ and $\mathbf{F}_m$, and apply a $1\times1$ convolution followed by softmax to obtain a gate $\mathbf{G}_m$. 
The gate adaptively balances general and specific features to obtain fused feature $\mathbf{F}_m^{F}$
\begin{equation}
\mathbf{F}_m^{F}=\mathbf{G}_m \odot \mathbf{F}_m^G + (1-\mathbf{G}_m) \odot \mathbf{F}_m,
\end{equation}
enabling the $\mathbf{F}_m^{F}$ to remain robust to disturbances in flood image patches while preserving fine-grained flood details.
The $\mathbf{F}_m^{F}$ are then input to the flood mask decoder to obtain the flood segmentation results. As depicted in Fig.~\ref{fig:pixel_deco}(d), the hierarchical fused features are passed through multiple decoder blocks and upsampling operations, ultimately producing the segmentation results.

\begin{table}[t]
\centering
\setlength{\belowcaptionskip}{-9pt}
\small
\renewcommand{\arraystretch}{1.05}
\setlength{\tabcolsep}{4pt}
\begin{tabular}{
  >{\centering\arraybackslash}m{3.7cm}
  >{\centering\arraybackslash}m{1.5cm}
  >{\centering\arraybackslash}m{1.1cm}
  >{\centering\arraybackslash}m{1.1cm}
}
\toprule
\addlinespace[5pt]
\textbf{Models} & \textbf{ROUGE-L} & \textbf{CIDER} & \textbf{CLAIR} \\
\addlinespace[3pt]
\midrule
Kimi-2.6 \cite{kimi} & 0.2526 & 0.1056 & 0.5599 \\
Doubao-Seed-2.0-Lite \cite{doubao} & 0.2449 & 0.0706 & 0.4097 \\
Doubao-Seed-2.0-Pro \cite{doubao} & 0.2653 & 0.0924 & 0.4498 \\
Doubao-Seed-2.1-Turbo \cite{doubao} & 0.2696 & 0.0929 & 0.3945 \\
GLM-5.3-Flash \cite{glm5} & 0.2590 & 0.0888 & 0.5209 \\
DeepSeek-V4-Flash \cite{deepseekv4} & 0.2721 & 0.0967 & 0.4230 \\
Qwen3-VL-2B \cite{qwen3} & 0.2362 & 0.0601 & 0.3648 \\
Qwen3-VL-4B \cite{qwen3} & 0.2129 & 0.0666 & 0.3295 \\
Qwen3-VL-32B \cite{qwen3} & 0.2601 & 0.0962 & 0.5260 \\
Qwen3-VL-2B$^{\dagger}$ \cite{qwen3} & 0.5034 & 0.3009 & 0.6758 \\
Qwen3-VL-4B$^{\dagger}$ \cite{qwen3} & 0.5198 & 0.3135 & 0.7066 \\
\midrule
GeoPixel-7B \cite{geopixel} & 0.0972 & 0.0062 & 0.0538 \\
TeoChat-7B \cite{teochat} & 0.1578 & 0.0143 & 0.1572 \\
GeoPixel-7B$^{\dagger}$ \cite{geopixel} & 0.3407 & 0.1478 & 0.3244 \\
TeoChat-7B$^{\dagger}$ \cite{teochat} & 0.5333 & 0.3233 & 0.7078 \\
\midrule
FloodVLM-Pix-2B$^{\dagger}$ & \textcolor{red}{0.5369} & \textcolor{red}{0.3328} & \textcolor{red}{0.7445} \\
FloodVLM-Pix-4B$^{\dagger}$ & \textcolor{blue}{\textbf{0.5390}} & \textcolor{blue}{\textbf{0.3381}} & \textcolor{blue}{\textbf{0.7586}} \\
\bottomrule
\end{tabular}
\caption{Language-level flood understanding results of SOTA methods and FloodVLM-Pix. $^{\dagger}$ denotes models fine-tuned on our dataset and models without $^{\dagger}$ are evaluated by direct inference. The best result in each column is shown in \textcolor{blue}{\textbf{blue}}, and the second-best is shown in \textcolor{red}{red}.}
\label{tab:language_results}
\end{table}

\section{Experiment}

\subsection{Implementation Details}

We adopt LSNet-S \cite{lsnet} as the backbone of the prior module. The prior module is trained for 100 epochs with a batch size of 128 using the AdamW \cite{adamw} optimizer with a learning rate of $1 \times 10^{-3}$, weight decay of $0.025$, $\beta=(0.9, 0.999)$, and a cosine annealing schedule with a minimum learning rate of $1 \times 10^{-5}$. Then, we leverage Qwen3-VL-2B/4B \cite{qwen3} as the backbone for the vision encoder, projector, and the LLM. The LLM is fine-tuned using LoRA \cite{lora} with a rank of 128, and the projection layer is fully fine-tuned. The vision encoder remains frozen. Fine-tuning is performed with a learning rate of $2 \times 10^{-5}$ for one epoch. Simultaneously with fine-tuning the LLM and projector, we train the pixel decoder with ConvNeXt \cite{convnet} as the backbone for 100 epochs with a learning rate of $1 \times 10^{-4}$. The entire FloodVLM-Pix model is trained on a single NVIDIA A100-80GB GPU.


\subsection{Baselines}

To rigorously evaluate the flood language understanding capability of FloodVLM-Pix, we establish two groups of baselines, i.e., general-domain VLMs and remote sensing VLMs. For general-domain VLMs, we evaluate several state-of-the-art (SOTA) models on the FloodVL-Pix dataset, including Kimi-2.6 \cite{kimi}, Doubao-Seed-2.0 (lite and pro), Doubao-Seed-2.1-turbo \cite{doubao}, GLM-5.3-Flash \cite{glm5}, DeepSeek-V4-Flash \cite{deepseekv4}, and Qwen3-VL variants (2B, 4B, and 32B) \cite{qwen3}, which represent advanced capabilities in general visual task processing. For remote sensing VLMs, we compare against GeoPixel-7B \cite{geopixel} and TeoChat-7B \cite{teochat}. GeoPixel-7B is an advanced remote sensing VLM that supports language understanding of fine-grained objects in remote sensing imagery. TeoChat-7B is a SOTA multi-temporal remote sensing VLM capable of interpreting temporal changes in remote sensing imagery. 

\begin{figure*}[!t]
    \centering
    \includegraphics[width=\textwidth]{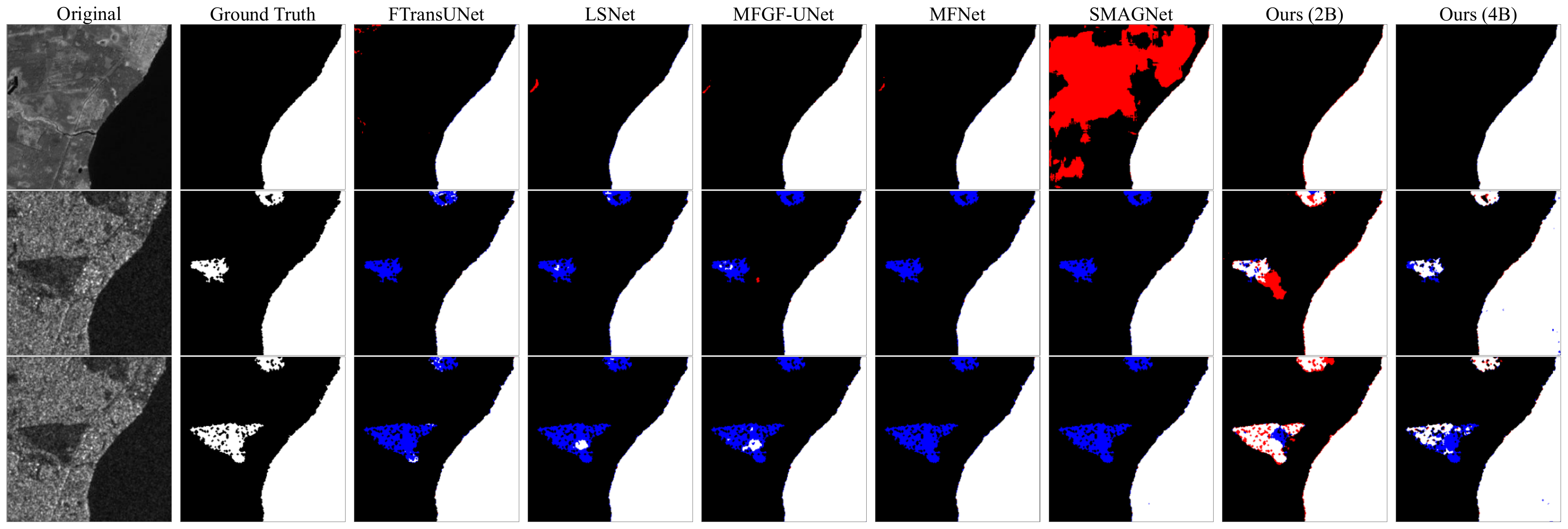}
    \setlength{\abovecaptionskip}{-6pt}
    \setlength{\belowcaptionskip}{-8pt}
    \caption{Results of the SOTA flood segmentation models and our proposed models. The white, red, blue, and black pixels are the regions of true positives, false alarms, missed targets, and true negatives, respectively. Our method yields fewer missed detections, demonstrating the effectiveness of the proposed pixel decoder.}  
    \label{fig:seg_result}
\end{figure*}

To accurately assess the flood segmentation capability of FloodVLM-Pix, we reproduce several SOTA flood segmentation methods, i.e., FTransUNet \cite{ftransunet}, MFGF-UNet \cite{mfgfunet}, MFNet \cite{mfnet}, and SMAGNet \cite{smagnet}. LSNet \cite{lsnet}, a SOTA backbone, is also trained on the FloodVL-Pix to serve as a strong segmentation baseline.

\subsection{Flood Description and Segmentation Results}

\begin{table}[t]
\centering
\setlength{\belowcaptionskip}{-12pt}
\small
\renewcommand{\arraystretch}{1.05}
\setlength{\tabcolsep}{3.5pt}
\begin{tabular}{lccccc}
\toprule
\noalign{\vspace{-0pt}}
\makecell[c]{\textbf{Model}} &
\makecell[c]{\textbf{Precision}} &
\makecell[c]{\textbf{Recall}} &
\makecell[c]{\textbf{OA}} &
\makecell[c]{\textbf{F1}} &
\makecell[c]{\textbf{IoU}} \\ [-0pt]
\midrule
FTransUNet \cite{ftransunet} & 0.6524 & 0.9583 & 0.8921 & 0.7763 & 0.6344 \\
LSNet \cite{lsnet} & \textcolor{blue}{\textbf{0.9156}} & 0.8191 & 0.9557 & 0.8647 & 0.7616 \\
MFGF-UNet \cite{mfgfunet} & 0.8134 & 0.9294 & 0.9510 & 0.8675 & 0.7661 \\
MFNet \cite{mfnet} & 0.7514 & \textcolor{red}{0.9600} & 0.9301 & 0.8430 & 0.7286 \\
SMAGNet \cite{smagnet} & 0.7679 & \textcolor{blue}{\textbf{0.9622}} & 0.9358 & 0.8541 & 0.7454 \\
\midrule
FloodVLM-Pix-2B & \textcolor{red}{0.8407} & 0.9497 & \textcolor{blue}{\textbf{0.9606}} & \textcolor{blue}{\textbf{0.8919}} & \textcolor{blue}{\textbf{0.8049}} \\
FloodVLM-Pix-4B & 0.8367 & 0.9266 & \textcolor{red}{0.9565} & \textcolor{red}{0.8793} & \textcolor{red}{0.7847} \\
\bottomrule
\end{tabular}
\setlength{\belowcaptionskip}{-8pt}
\caption{Flood segmentation results of existing SOTA methods and FloodVLM-Pix. The best and second best results are in \textcolor{blue}{\textbf{blue}} and \textcolor{red}{red}, respectively. Our FloodVLM-Pix-2B/4B achieve higher overall segmentation performance, i.e., OA, F1, IoU. }
\label{tab:segmentation_results}
\end{table}

\textbf{Flood description results.} In Table \ref{tab:language_results}, we compare the flood language-level understanding capability of FloodVLM-Pix with SOTA general-domain and remote sensing VLMs. FloodVLM-Pix-2B(-4B) adopts Qwen3-VL-2B(-4B) as its backbone. All models are evaluated using a comprehensive set of metrics, i.e., \textit{recall-oriented understudy for gisting evaluation} (ROUGE-L) \cite{rouge}, \textit{consensus-based image description evaluation} (CIDER) \cite{cider}, and \textit{criterion using language models for image caption rating} (CLAIR) \cite{clair}. For all these metrics, higher scores indicate better performance in flood language description. Additional metrics and explanations for evaluating the flood description capability of the VLMs can be found in the supplementary material. Our FloodVLM-Pix-4B achieves SOTA performance on all metrics because we incorporate a prior module to guide the VLM toward accurate flood language understanding. The direct inference models (i.e., the models without $^{\dagger}$) in Table \ref{tab:language_results} show poor performance, stemming from limited SAR interpretation ability and attentional bias of VLMs during flood language understanding. Table \ref{tab:language_results} also reports results of models fine-tuned on our FloodVL-Pix dataset (Qwen3-VL-2B$^{\dagger}$/4B$^{\dagger}$, TeoChat-7B$^{\dagger}$, and GeoPixel-7B$^{\dagger}$), and these models also underperform FloodVLM-Pix because they lack designs to address the attentional bias for large-scale floods.

\textbf{Flood Segmentation results.} Table \ref{tab:segmentation_results} shows the results of flood segmentation. We evaluate FloodVLM-Pix and SOTA methods on \textit{intersection over union} (IoU) \cite{iou}, \textit{F1-score} (F1) \cite{f1_pre_rec}, \textit{overall accuracy} (OA) \cite{OA}, \textit{precision} \cite{f1_pre_rec}, and \textit{recall} \cite{f1_pre_rec}. Although LSNet \cite{lsnet} achieves high precision, it does so at the cost of low recall (i.e., many missed detections). Conversely, MFNet \cite{mfnet} and SMAGNet \cite{smagnet} obtain high recall values but suffer from high false alarms (low precision). Consequently, LSNet, MFNet, and SMAGNet suffer from low overall flood segmentation performance in terms of OA, F1, and IoU values. Our FloodVLM-Pix employs a gate to adaptively fuse general flood contextual features with single-image-specific features, achieving high overall segmentation performance. We also present visualization results of flood segmentation in Fig.~\ref{fig:seg_result}, where our models yield fewer missed detections, demonstrating the effectiveness of the proposed pixel decoder.

\subsection{Ablation Experiments}

\begin{figure}[!t]
    \centering
    \includegraphics[width=\columnwidth]{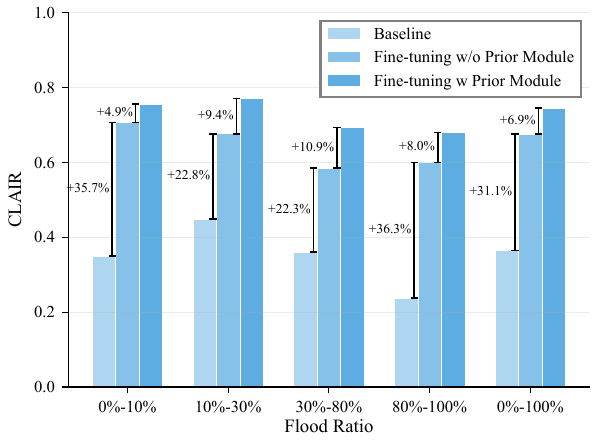}
    \setlength{\abovecaptionskip}{-10pt}
    \setlength{\belowcaptionskip}{-12pt}
    \caption{Flood description CLAIR of direct inference (baseline), fine-tuning with/without the prior module on Qwen3-VL-2B. Fine-tuning refers to LoRA-based fine-tuning \cite{lora} of the LLM and full fine-tuning of the projection layer, while the prior module is kept frozen. ``w''/``w/o'' prior module denotes fine-tuning with and without the flood prior information output by the prior module, respectively. Flood ratio is the proportion of the flood area in the image patch, and the last column (0\%-100\%) summarizes performance across all flood ratios, equating to the average CLAIR. Fine-tuning substantially improves performance, and incorporating the prior module further yields an average gain of 6.9\% in CLAIR.}  
    \label{fig:prior_module}  
\end{figure}


\textbf{Effect of Prior Module.} We evaluate the flood language understanding performance of direct inference (baseline) and fine-tuning with/without the prior module on Qwen3-VL-2B, as shown in Fig. \ref{fig:prior_module}. Fine-tuning includes LoRA-based fine-tuning \cite{lora} of the LLM and full fine-tuning of the projection layer, while the prior module is kept frozen, as depicted in Section 4.2. ``With/without prior module'' denotes fine-tuning with/without the flood prior information output by the prior module. We draw three main conclusions. 1) Baseline performs poorly overall, especially when the flood ratio is high ($>80\%$). This is because when floods occupy a large proportion of the image patch, they are easily misclassified as background. 2) Fine-tuning (without the prior module) on our dataset brings substantial gains, with an average improvement of 31.1\% in CLAIR, demonstrating that the VLM acquires stronger flood understanding capability than the baseline. For the high-flood-ratio cases ($>80\%$), the improvement of CLAIR reaches 36.3\%, validating that fine-tuning significantly mitigates the tendency of the model to misclassify large-scale floods as background. 3) Incorporating the prior module yields an average gain of 6.9\% in CLAIR. 
In the case of large-scale floods ($>80\%$), adding the prior module improves CLAIR by 8.0\%, indicating that accurate flood priors recalibrate attention from small-scale objects toward flooded regions and supply explicit region-wise spatial information for severity and evolution-aware flood understanding.
In the intermediate flood-ratio range (10\%-80\%), the improvements after using the prior module are also significant. In this range, floods often appear only in specific regions (e.g., the upper-left corner), and an accurate flood prior substantially helps the model understand where to attend, thereby improving the description performance of flood severity and evolution.



\begin{table}[t]
\centering
\small
\renewcommand{\arraystretch}{1.05}
\setlength{\tabcolsep}{3.5pt}
\begin{tabular}{ccccc}
\toprule
\noalign{\vspace{-0pt}}
\makecell[c]{\textbf{Model}} &
\makecell[c]{\textbf{OA}} &
\makecell[c]{\textbf{F1}} &
\makecell[c]{\textbf{IoU}} \\ [-0pt]
\midrule
w/o LN + Add   & 0.9447 & 0.8525 & 0.7429 \\
LN + Add   & 0.9539 & 0.8764 & 0.7800 \\
LN + Concat   & \textcolor{blue}{\textbf{0.9606}} & \textcolor{blue}{\textbf{0.8919}} & \textcolor{blue}{\textbf{0.8049}} \\
w/o General-Specific Fusion Module  & \textcolor{red}{0.9564} & \textcolor{red}{0.8823} & \textcolor{red}{0.7894} \\
w/o Features from Qwen & 0.9556 & 0.8799 & 0.7856 \\
\bottomrule
\end{tabular}
\setlength{\belowcaptionskip}{-8pt}
\caption{Ablation experiments of different pixel decoders. The "w/o LN"/"LN" means the Qwen feature extraction architecture in Fig. \ref{fig:pixel_deco}(b) without/with layer normalization layers. ``Add''/``Concat'' denotes that either element-wise addition or channel-wise concatenation is employed to fuse features from the flood image patch and those from Qwen in Fig. \ref{fig:pixel_deco}(b). The best and second best results are in \textcolor{blue}{\textbf{blue}} and \textcolor{red}{red}, respectively.}
\label{tab:pixel_deco}
\end{table}

\textbf{Architecture of the Pixel Decoder.} We evaluate different architectures of pixel decoders and use three metrics, i.e., OA, F1, and IoU, to assess the flood segmentation capability of these pixel decoders, as reported in Table~\ref{tab:pixel_deco}. First, we remove the layer normalization (LN) layers in the Qwen feature extractor (four upsampling + convolution + LN operations) in Fig.~\ref{fig:pixel_deco}(b). This leads to inferior flood segmentation performance, as the absence of LN causes unstable optimization and degraded feature representations, which directly impair the ability of the decoder to distinguish floods from background and thus lower the OA, F1, and IoU. Hence, LN is adopted throughout.
We further replace the channel-wise concatenation in Fig.~\ref{fig:pixel_deco}(b) with element-wise addition and compare the results to investigate how different fusion strategies between features from the Qwen3 vision encoder and the original image patch affect flood segmentation performance.
Concatenation yields better results, as it preserves full information and enables subsequent layers to selectively integrate information from both feature streams. In contrast, addition incurs information loss, degrading flood segmentation performance. Finally, we remove the general-specific feature fusion module or the features from the Qwen3 vision encoder entirely, both of which degrade flood segmentation performance, confirming the effectiveness of both components.

\section{Conclusion}


We construct FloodVL-Pix, the first flood benchmark for language and pixel-level, multi-modal, multi-temporal flood understanding, advancing disaster-oriented Earth observation and flood response. We propose FloodVLM-Pix, which integrates a flood prior module and a multi-temporal flood pixel decoder to achieve accurate flood understanding. The flood prior module provides accurate priors to mitigate the attentional bias of VLMs in flood language understanding. The pixel decoder fuses general and specific features in a gated manner to aggregate cross-modal and cross-temporal flood context while preserving image-specific details, enabling accurate multi-temporal flood segmentation under various disturbances. Experiments show that FloodVLM-Pix outperforms SOTA methods in both flood description and segmentation.

\section{Acknowledgment}
This work was supported by National Key Research and Development Program of China under Grant 2025YFF0514602, Beijing Natural Science Foundation under Grant JQ25013, and National Natural Science Foundation of China under Grants  62622123, 62471274 and 62501359.


{
    \small
    \bibliographystyle{ieeenat_fullname}
    \bibliography{main}
}

\end{document}